\documentclass{article}

\usepackage[numbers]{natbib}

\usepackage[preprint]{neurips_2024}

\usepackage[utf8]{inputenc} 
\usepackage[T1]{fontenc}    
\usepackage{hyperref}       
\newcommand\nnfootnote[1]{%
  \begin{NoHyper}
  \renewcommand\thefootnote{}\footnote{#1}%
  \addtocounter{footnote}{-1}%
  \end{NoHyper}
}
\usepackage{url}            
\usepackage{booktabs}       
\usepackage{amsfonts}       
\usepackage{nicefrac}       
\usepackage{microtype}      
\usepackage[dvipsnames]{xcolor}
\usepackage{subfigure}
\usepackage{xspace}

\usepackage{amsmath,amsfonts,amssymb,amsthm}
\usepackage{pifont}
\usepackage{soul}           
\usepackage{appendix}

\usepackage{color}
\usepackage{graphicx}
\usepackage{mathrsfs}
\usepackage{subcaption}
\usepackage{makecell}
\usepackage{algorithm,bbm}
\usepackage{algorithmic}
\usepackage{multirow}
\usepackage{lineno}
\usepackage{enumerate}

\DeclareMathOperator{\QL}{QL}
\DeclareMathOperator{\WQE}{WQL}

\def\benu{\begin{enumerate}}
\def\eenu{\end{enumerate}}

\newcommand\btheta{\boldsymbol \theta}

\newcommand{\model}[1]{\text{\texttt{#1}}\xspace}

\newcommand{\MQCNN}{\model{MQCNN}}
\newcommand{\MQT}{\model{MQT}}

\newcommand{\PeakAttention}{\model{PeakAttention}}

\newcommand{\ADAM}{\model{ADAM}}

\newcommand{\ours}{\model{SPADE}}

\newcommand{\first}[1]{\textbf{#1}}

\newcommand{\dataset}[1]{\texttt{#1}}

\newcommand{\TourismL}{\dataset{Tourism-L}}

\newcommand{\PEGains}{3.9\%\xspace}
\newcommand{\PPEGains}{4.5\%\xspace}

\title{$\spadesuit$ \ours $\spadesuit$\\Split Peak Attention DEcomposition}

\author{%
  Malcolm Wolff \textsuperscript{*} \\
  Amazon \\
  \texttt{wolfmalc@} \\
  \And
  Kin G.\ Olivares\textsuperscript{*} \\
  Amazon \\
  \texttt{kigutie@} \\
  \And
  Boris Oreshkin \\
  Amazon \\
  \texttt{oreshkin@} \\
  \And
  Sunny Ruan \\
  Amazon \\
  \texttt{jruan@} \\
  \And
  Sitan Yang \\
  Amazon \\
  \texttt{sitanyan@} \\
  \And
  Abhinav Katoch \\
  Amazon \\
  \texttt{abkatoch@} \\
  \And
  Shankar Ramasubramanian \\
  Amazon \\
  \texttt{sramasub@} \\
  \And
  Youxin Zhang \\
  Amazon \\
  \texttt{youxz@} \\
  \AND
  Michael W.\ Mahoney \\
  Amazon \\
  \texttt{zmahmich@} \\
  \And
  Dmitry Efimov \\
  Amazon \\
  \texttt{defimov@} \\
  \And
  Vincent Quenneville-Bélair \\
  Amazon \\
  \texttt{quennv@} \\
}

\begin{document}

\maketitle
\nnfootnote{$^*$ These authors contributed equally.}

\vspace{-0.5cm}
\begin{abstract}
    Demand forecasting faces challenges induced by Peak Events (PEs) corresponding to special periods such as promotions and holidays. 
    Peak events create significant spikes in demand followed by demand ramp down periods. 
    Neural networks like \MQCNN~\citep{wen2017mqrcnn,olivares2021pmm_cnn} and \MQT~\citep{eisenach2020mqtransformer} overreact to demand peaks by carrying over the elevated PE demand into subsequent Post-Peak-Event (PPE) periods, resulting in significantly over-biased forecasts. 
    To tackle this challenge, we introduce a neural forecasting model called Split Peak Attention DEcomposition, \ours. 
    This model reduces the impact of PEs on subsequent forecasts by modeling forecasting as consisting of two separate tasks: one for PEs; and the other for the rest. 
    Its architecture then uses masked convolution filters and a specialized Peak Attention module. We show \ours's performance on a worldwide retail dataset with hundreds of millions of products. Our results reveal an overall PPE improvement of \PPEGains, a 30\% improvement for most affected forecasts after promotions and holidays, and an improvement in PE accuracy by \PEGains, relative to current production models.
\end{abstract}

\section{Introduction}

Forecasting methods based on neural networks have produced accuracy improvements across multiple forecasting application domains such as large e-commerce retail~\citep{wen2017mqrcnn,eisenach2020mqtransformer,olivares2021pmm_cnn,Quenneville-Belair2023}, financial trading~\citep{sezer2020financial}, planning and transportation~\citep{laptev2018transfer_regloss}, and forecasting competitions~\citep{smyl2019esrnn,oreshkin2020nbeats}.
For large e-commerce, neural networks such as \MQCNN~\citep{wen2017mqrcnn,olivares2021pmm_cnn} and \MQT~\citep{eisenach2020mqtransformer} forecast product demand, based on product demand history, product attributes and known future information (e.g.,\ promotions, planned sales, holidays), etc.
Product demand frequently exhibits peaks caused by the impact of Peak Events (PEs) such as promotions, deals, or holiday sales; forecast is particularly important during and after these peaks since we are more likely to run into inventory constraints.

The \emph{carry-over} effect manifests itself as significant over-bias in the forecasts following demand spikes characteristic of PEs. Forecasting errors during and after PEs lead to logistical challenges: increased storage and operational costs, and often necessitate manual interventions to adjust forecasts, compromising the link between our provided forecast and downstream decisions.
To handle this effect, filtering techniques such as \citep{muler2009robust_arma, crevits2016robust_ets} have been used in traditional time series analysis.
In this paper, we close the gap between such techniques and neural network based forecasts by developing a new forecasting architecture called \ours capable of maintaining accuracy in the presence of PEs.

Our key contributions are summarized below:

\begin{enumerate}[(i)]
    \item \textbf{Masked Convolutions.}
    We decompose the historical time series features into peak and non-peak components, based on \emph{a priori} causal indicators. The convolutional encoder experiences only the historical demand without peaks.
    \item \textbf{Peak Attention.}
    The peaks are encoded by a specialized attention mechanism we call Peak Attention. This module leverages \emph{a priori} causal information, allowing the forecast to react quickly during and after PEs.
    \item \textbf{Accuracy Improvements.}
    We expand upon the experiments' dataset size of \MQCNN and \MQT in \citep{wen2017mqrcnn, eisenach2020mqtransformer} from millions to hundreds of millions of series, and show post PE accuracy improvements of \PPEGains, and PE accuracy improvements of \PEGains over \MQCNN and \MQT respectively.
\end{enumerate}

We organize the rest of the paper as follows.
We introduce our proposed \ours method in Section~\ref{section3:methodology}.
Section~\ref{section4:evaluation} contains main experiments and ablation studies. We summarize the results and mention future research directions in Section~\ref{section5:conclusion}.
 \label{section1:introduction}
\section{\ours}
\label{section3:methodology}

\begin{figure}[t] 
    \centering
    \includegraphics[width=120mm]{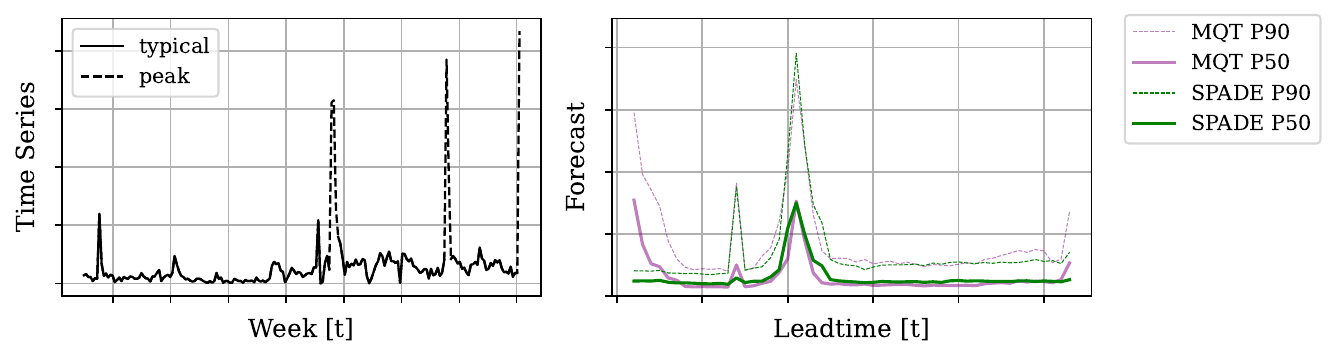} 
    \caption{Illustration of ``carry-over'' degradation, visible in \MQT's forecast downward trend. Peak values carry-over, degrading \MQT's forecast accuracy, whereas \ours does not exhibit such an effect.
    }
    \label{fig:motivation2}
\end{figure}

Accurate forecasting requires integrating multiple inputs from other time series and data modalities, such as static information or exogenous factors, rather than relying solely on historical patterns. Additionally, effective models must jointly optimize across multiple tasks, considering not just standard metrics like MAE (Mean Absolute Error), but also the broader distribution of forecasts, or across different time granularities. Finally, robust forecasting must reliably capture both long-term trends and sharp peak events to be effective in practical settings.



\ours is a Sequence to Sequence forecasting architecture~\citep{sutskever2014seq2seq_translation} initially built for the purpose of product demand forecasting which integrates past time series data $\mathbf{x}^{(p)}_{[t]}$, static information $\mathbf{x}^{(s)}$ (e.g., product text features), and known future information $\mathbf{x}^{(f)}_{[t][h]}$ (e.g., world event indicators, countries and seller promotions). The model simultaneously predicts outcomes across multiple quantiles and levels of aggregation—each with unique patterns and noise characteristics—and leverages \emph{a priori} event information to split Peak Events (PEs) from the remainder of the time series, eliminating ``carry-over'' effects many current time series forecasting models suffer from (see Figure~\ref{fig:motivation2}).

\begin{figure}[t]
\centering
\includegraphics[scale=.20]{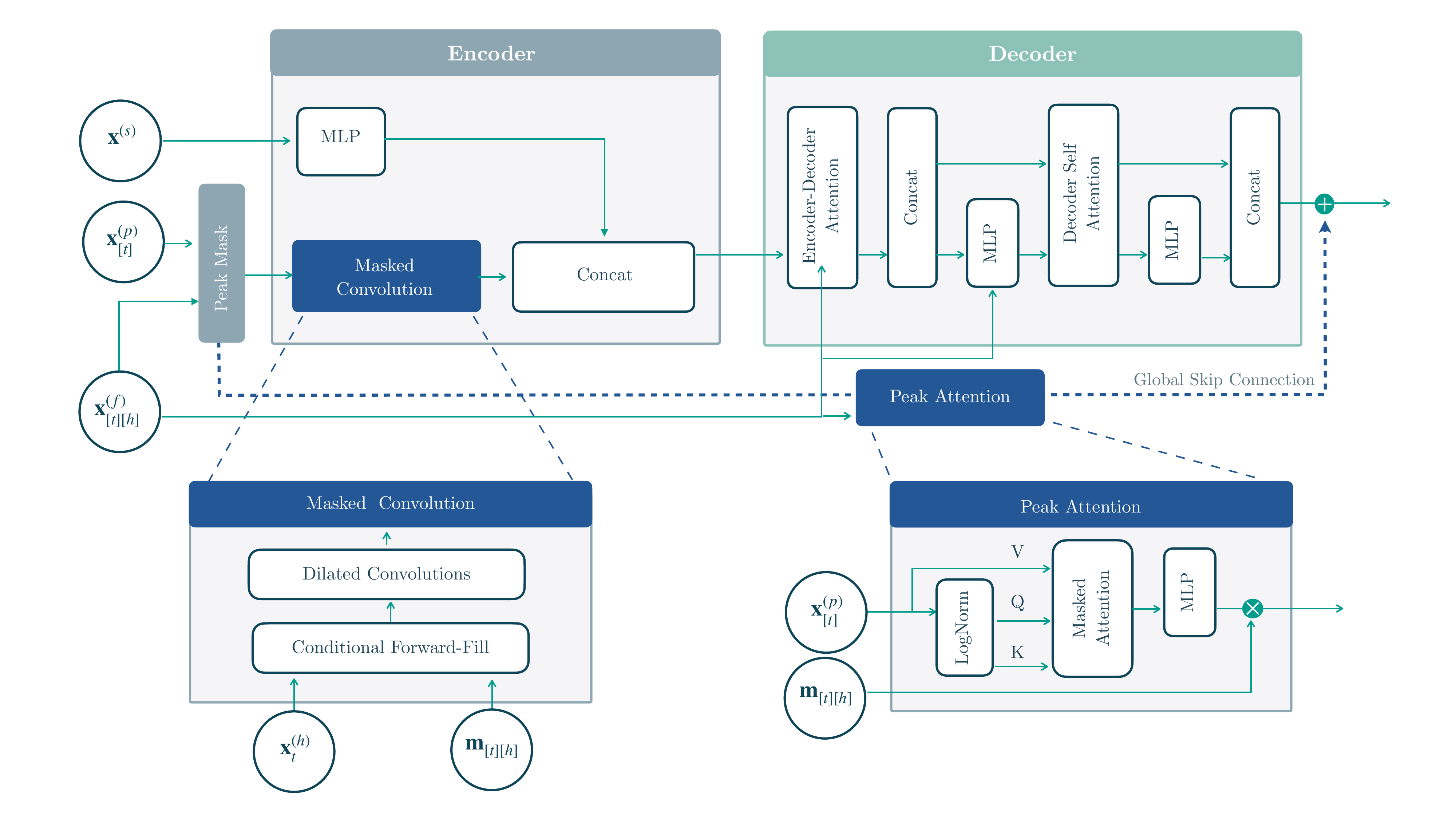}
\caption{\ours decomposes its temporal features to distinguish usual behavior from peaks.}
\label{fig:splitformer-arch}
\end{figure}

Figure~\ref{fig:splitformer-arch} displays the architecture.
The \texttt{PeakMask} generates PE indicators from known future information, creating a mask to decompose the time series into PEs and non-PEs. The \texttt{RobustConvolution} block filters PEs from the historic inputs with a forward-fill operation and encodes the result with a series of dilated convolutions.
The \texttt{PeakAttention} block forecasts PE magnitudes using past PE information and known future information, and the PE and non-PE partial predictions are summed to produce the final forecast.
Architecture details are provided in Appendix~\ref{appendix:arch_details}.

%

Forecast accuracy is evaluated with the weighted quantile loss (WQL)
\begin{align}
\label{equation:WQL}
\mathrm{WQL}(\mathbf{y}, \hat{\mathbf{y}}^{(q)}; q, \mathcal{I}, \mathcal{H}) &= \frac{\sum_{i} \sum_{t} \sum_{h}   \QL\left(y_{i,t,h},\; \hat{y}^{(q)}_{i,t,h}(\btheta) ; q\right) } {\sum_{i} \sum_{t} \sum_{h}   y_{i,t,h} } ,
\end{align}
where $\QL\left(y,\; \hat{y}^{(q)}; q\right)=q(y-\hat{y})_{+}+(1-q)(\hat{y}-y)_{+}$ is the quantile loss function, $\hat{y}^{(q)}$ denotes the estimated quantile, $\btheta$ denotes a model in the class of models $\Theta$ defined by the model architecture. We optimize $\btheta$ by minimizing the numerator of equation~\eqref{equation:WQL} summed across the quantiles of interest.
See Appendix~\ref{appendix:training_methodology} for details.
In addition to WQL, we evaluate the WQL during PEs and Post-PE (denoted PPE) to capture accuracy at PEs and the ``carry-over'' effect, respectively; letting $\mathcal{I}_{\text{PE}} \equiv \{i \in \mathcal{I}\ \mid\ d_{i,t} = 1\}$ where $d_{i,t}$ represents time series deal information, we have
\begin{align}
\WQE_{\text{PE\phantom{P}}} &= \WQE(\mathbf{y}, \hat{\mathbf{y}}^{(q)} ; q, \mathcal{I}_{\text{PE}}, \mathcal{H}_{\text{PE}}), \quad\text{and}\quad
\WQE_{\text{PPE}} &= \WQE(\mathbf{y}, \hat{\mathbf{y}}^{(q)} ; q, \mathcal{I}, \mathcal{H}_{\text{PPE}})  ,
\end{align}
where $\mathcal{H}_{\text{PE}}$ are the horizons including a PE occur, and $\mathcal{H}_{\text{PPE}}$ are the horizons after.
A typical demand pattern indicating seasonality, PE demand peak and PPE carry-over effect is shown in Figure~\ref{fig:motivation2}.

\begin{table*}[t]
\footnotesize
\centering
\caption{Empirical evaluation of the weighted quantile loss (WQL). We evaluate P50 and P90 WQL overall, during and after PEs.
The best result is shown in bold, lower is better. \textsuperscript{*} We report the PPE WQL of the most affected series after holidays and promotions.
}

\label{table:empirical_evaluation}
\begin{tabular}{l|llll} \toprule
    Metric                                            & \ours          & \MQCNN   & \MQT   \\ \midrule
    P50 WQL                                           & \first{0.9912} & 1.1842   & 1.0000 \\
    P90 WQL                                           & \first{0.9935} & 1.1990   & 1.0000 \\
    $\text{P50 WQL}_{\text{PE}}$                      & \first{0.9672} & 0.9740   & 1.0000 \\
    $\text{P90 WQL}_{\text{PE}}$                      & \first{0.9557} & 1.0068   & 1.0000 \\
    \textsuperscript{*}$\text{P50 WQL}_{\text{PPE}}$  & \first{0.7012} & 0.9386   & 1.0000 \\
    \textsuperscript{*}$\text{P90 WQL}_{\text{PPE}}$  & \first{0.7727} & 0.9380   & 1.0000 \\ \bottomrule
\end{tabular}
\end{table*}

\section{Experiments}
\label{section4:evaluation}

The dataset consists of hundreds of millions of series across several countries covering three years training data (2019-2022) and one year for evaluation (2023).
Table~\ref{table:empirical_evaluation} displays results of empirical evaluation for \ours, \MQCNN\, and \MQT\ across countries. \MQCNN\ and \MQT\ show a distinct trade-off surrounding PEs. Both \MQT and \ours improve by 18\% and 19\% the overall P50/P90 WQL when compared to its predecessor the \MQCNN. However, \MQT PE WQL has not a significant improvement compared to \MQCNN, we believe this is due to optimization overfit and P90 undercalibration.

\MQT's P50/P90 PPE WQL degrades compared to \MQCNN by 7\%, we are inclined to believe that the peak carry-over effect of the encoded embeddings is higher than \MQCNN's for this dataset. \ours improves upon \MQCNN and \MQT by 30\% in P50 PPE WQL and 23\% P90 PPE WQL for most affected series after holidays and promotions. The overall PPE P50/P90 improvements of \ours compared to \MQT are of 4.5\%.

\ours's specialized attention to peak events result in large improvements. \ours\ outperforms \MQT by 3\% on P50 WQL and 4\% on P90 WQL during peaks.

To further clarify and analyze the source of improvements we performed an ablation study on variants of the \MQCNN and \MQT, for a smaller dataset in Appendix~\ref{appendix:spade_ablation}.

\begin{figure}[t] 
    \centering
    \includegraphics[width=0.48\linewidth]{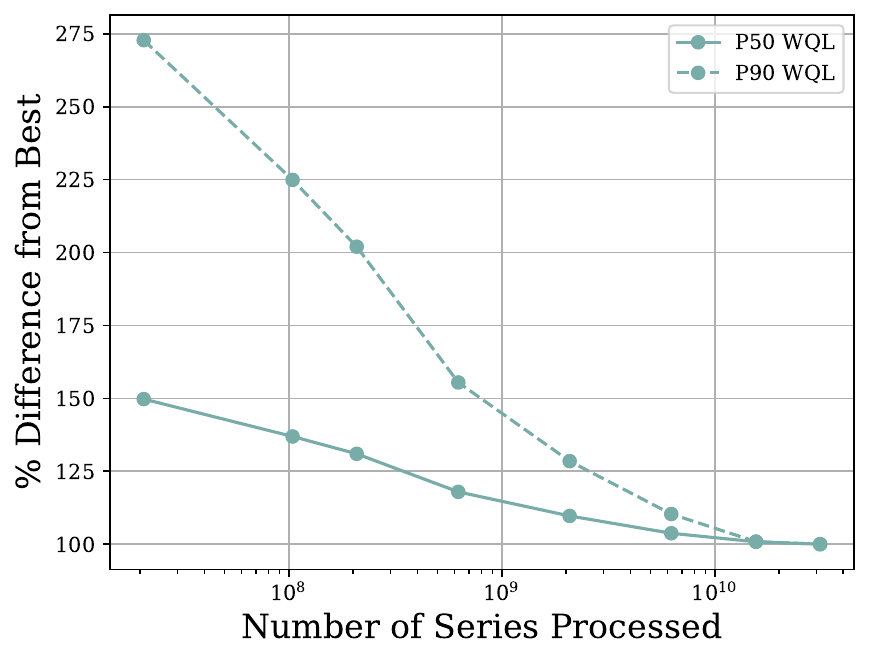} 
    \caption{\ours\ shows evidence of forecast accuracy scaling with training time series.}
    \label{fig:scaling}
\end{figure}

Finally, we study the scaling properties of our model. Our results, represented by Figure~\ref{fig:scaling}, show that forecast accuracy scales with the number of series during training, with P90 WQL improving more substantially than P50 WQL.
\section{Conclusions}

We introduced a novel forecasting method, \ours, that combines two complementary techniques (splitting temporal features into peak and non-peak components before encoding them, and a specialized peak attention module to enhance forecast sharpness) to identify and deal with spikes in realistic forecasting systems. 
\ours  alleviates the peak \emph{carry-over} effect, showing higher accuracy during and after the peaks events. 
In our retail demand forecasting experiments, we observe a PE accuracy improvement of \PEGains, and a reduction of the post peak degradation of \PPEGains, when compared to \MQCNN and \MQT production models. For the most affected series by promotions and holidays the improvements reached 30\%.


In our empirical evaluations, we applied peak masking convolutions to variants of \MQCNN and \MQT, but our \ours method is not restricted to convolutional encoders. 
Exploring its application to other recurrent encoders such as LSTMs, attention mechanisms, or MLPs presents a promising avenue for future research. 
Moreover, the current implementation of \ours relies on pre-existing causal and PE indicators, but there is potential to enhance the methodology by incorporating unsupervised detection techniques to identify PEs autonomously.

 \label{section5:conclusion}

\bibliographystyle{plain}
\bibliography{main.bib}

\clearpage
\appendix
\section{Architecture Details}

\begin{figure}[ht] 
    \centering
    \subfigure{
    \includegraphics[width=0.9\linewidth]{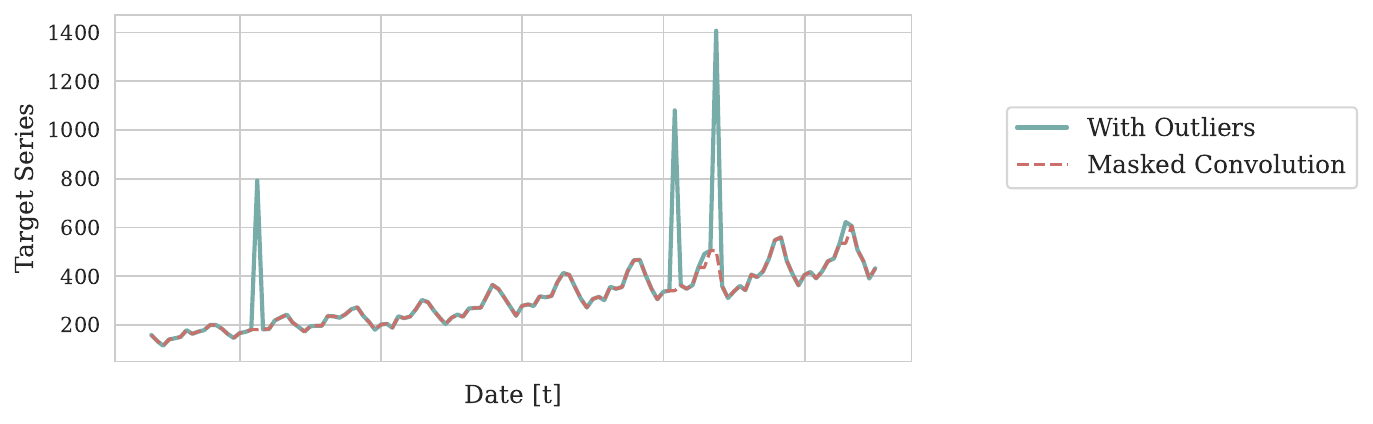} 
    }
    \caption{Masked convolutions enhance neural forecasting architectures by filtering peaks before inputting temporal features to the encoder, thus mitigating the peak carry-over effect.
    } 
    \label{fig:masked_convolution}
\end{figure}

\begin{figure}[h] 
    \centering
    \includegraphics[width=0.55\linewidth]{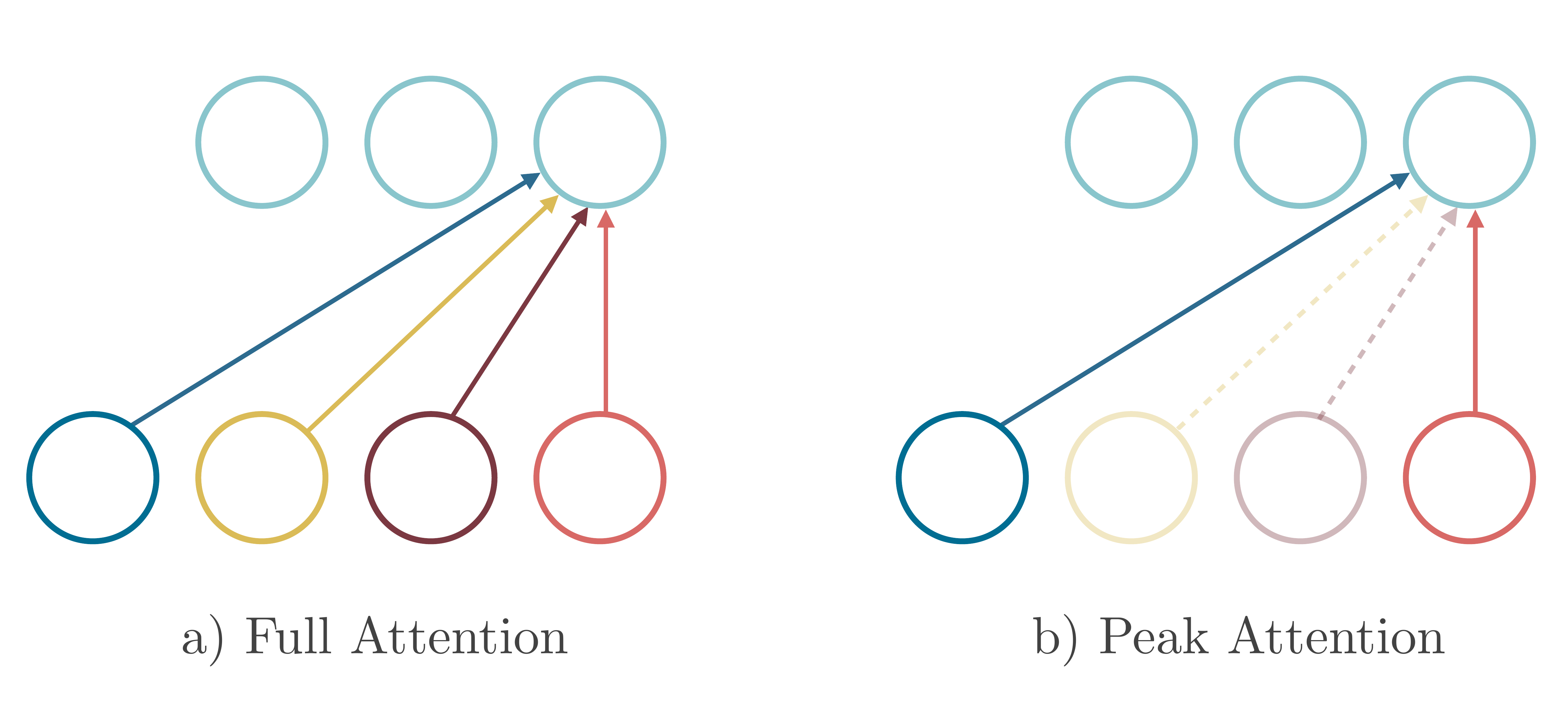} 
    \caption{The Peak Attention module regularizes the classic attention mechanism by sparsifying its weights using future covariate information.
    }
    \label{fig:peak_attention}
\end{figure}

\paragraph{Masked Convolution.} 
We use the causal indicators from the known future information $\mathbf{x}^{(f)}_{[t][h]}$, to create a mask that identifies peak temporal observations $\mathbf{m}_{[t][h]}$. 
The peak mask $\mathbf{m}_{[t][h]}$ is then used to decompose the time series into peak and non-peak observations. The $\mathbf{x}^{(p)}_{[t]}$ inputs of the convolutional encoder are thus filtered of their peak observations following:
\begin{align}
\mathbf{\tilde{x}}^{(p)}_{t}
&=
\begin{cases}
      \mathbf{x}^{(p)}_{t} & \mathrm{if} \quad \mathbf{m}_{t}=0
      \\
      \mathbf{x}^{(p)}_{t^{*}} & \mathrm{if} \quad \mathbf{m}_{t}=1,\; t^{*}=\min_{\tau\leq t}\mathbf{m}_{\tau}=0 
\end{cases}
\\
\mathbf{e}_{[t]}^{(p)} &= \text{Convolution}(\mathbf{\tilde{x}}^{(p)}_{[t]}) . \qquad\qquad\qquad
\end{align}

\paragraph{Peak Attention.}
This module computes a forecast update $\Delta_{[t][h]}$ using the peak mask $\mathbf{m}_{[t][h]}$, the past information $\mathbf{x}^{(p)}_{[t]}$ and known future information $\mathbf{x}^{(f)}_{[t][h]}$ to compute the following operations:
\begin{align}
    \mathbf{q}_{[t][h]} &= \mathrm{MLP}(\mathbf{e}^{(p)}_{[t]}, \mathbf{x}_{[t][h]}^{(f)})
    \quad
    \mathbf{k}_{[t][h]} = \mathrm{MLP}(\mathbf{e}_{[t]}^{(p)}, \mathbf{x}_{[t][h]}^{(f)})
    \quad
    \mathbf{v}_{[t][h]} = \mathrm{MLP}(\mathbf{e}_{[t]}^{(p)}, \mathbf{x}_{[t][h]}^{(f)})
    \\
    \mathbf{H}_{[t][h]} &= \mathrm{SoftMax}\left(\mathbf{q}_{[t][h]} \times \mathbf{k}^{\intercal}_{[t][h]} \times \mathbf{m}_{[t][h]}\right) \times \mathbf{v}_{[t][h]},\\
    \Delta_{[t][h]} &= \text{MLP}(\mathbf{H}_{[t]}) \times \mathbf{m}_{[t][h]}  ,
\end{align}
where $\mathbf{q}_{[t][h]}$, $\mathbf{k}_{[t][h]}$, and $\mathbf{v}_{[t][h]}$ are horizon-specific query, key, and value tensors.

The attention weights are masked to only use PEs from the past; in contrast to a complete horizon-time attention module, \PeakAttention requires only a fraction of the computation, as it only searches historical PEs instead of the full history to induce the forecasts' sharpness.
\begin{align}
    \mathbf{\hat{y}}_{[t][h]} = \Delta_{[t][h]}+\mathrm{MLPDecoder}\left(\mathbf{e}_{[t]}^{(p)}, \mathbf{e}^{(s)}\right),
\end{align}
where $\mathbf{e}_{[t]}^{(p)}$ are the encoded masked historic temporal features and $\mathbf{e}^{(s)}$ are the encoded static features.
We incorporate a global skip connection to adjust the predictions of the multi-horizon MLP decoder, effectively decomposing the forecasts into PEs and a baseline.
 \label{appendix:arch_details}
\section{Training Methodology and Hyperparameter Selection} \label{appendix:training_methodology}

\begin{table} [ht]
\caption{The \ours\ architecture parameters configured once. The second panel controls the optimized parameters, we only considered learning rates and training epochs. (\textsuperscript{*}The effective SGD batch is multiplied by the number of GPUs in the execution cluster.)}
\label{table:model_parameters}
\scriptsize
\centering
\begin{tabular}{lcccc}
    \toprule
    \textsc{Parameter}                                          & Notation          & \multicolumn{3}{c}{Considered Values} \\
                                                                &                   & \ours              & \MQT               & \MQCNN             \\ \midrule
    Single GPU SGD Batch Size\textsuperscript{*}               & -                 & 32 (10,240)        & 32 (10,240)        & 32 (10,240)        \\
    Main Activation Function                                  & -                 & ReLU               & ReLU               & ReLU               \\
    Max Temporal Convolution Kernel Size                       & -                 & 32                 & 32                 & 32                 \\
    Temporal Convolution Layers                                & -                 & 6                  & 6                  & 6                  \\ 
    Temporal Convolution Filters                               & -                 & 30                 & 30                 & 30                 \\
    Static Encoder D.Multip. $(\alpha\times|\sqrt{x}^{(s)}|)$  & -                 & 30                 & 30                 & 30                 \\
    Future Encoder Dimension (hf1)                             & -                 & 50                 & 50                 & 50                 \\
    Horizon Agnostic Decoder Dimensions                        & -                 & 100                & 100                & 100                \\
    Horizon Specific Decoder Dimensions                        & -                 & 20                 & 20                 & 20                 \\ 
    \PeakAttention Number of Heads                             & -                 & 4                  & 4                  & 4                  \\ \midrule
    Learning Rates                                             & -                 & $\{0.001,0.0001\}$ & $\{0.001,0.0001\}$ & $\{0.001,0.0001\}$ \\
    Number of Epochs                                           & -                 & $\{10, 20, 30\}$   & $\{10, 20, 30\}$   & $\{10, 20, 30\}$   \\ \bottomrule
\end{tabular}
\end{table}

\paragraph{Training Methodology.} Let $\btheta$ be a model that resides in the class of models $\Theta$ defined by the model architecture. Let $\mathcal{A}$ the dataset's products, and $\mathcal{H}$ the horizon defined by lead times and spans. We train a quantile regression model by minimizing the following multi-quantile loss:
\begin{align}
\label{equation:learning_objective}
    \min_{\btheta}
    \sum_{q} \sum_{i} \sum_{t} \sum_{h}  \QL\left(y_{i,t,h},\; \underline{\hat{y}}^{(q)}_{i,t,h}(\btheta) ;\ q\right),
\end{align}
for products $a\in \mathcal{A}$, time $t$ and horizon $h\in\mathcal{H}$, and $\underline{\hat{y}}^{(q)}$ denotes the estimated quantile\footnote{During training, demand and forecasts are normalized by the length of the horizon $h$.}. We optimize \ours using stochastic gradient descent with \emph{Adaptive Moments} (\ADAM; \cite{kingma2014adam_method}). 
Details on the implementation and optimization methodology and variants are available in Appendix~\ref{appendix:training_methodology}.


\paragraph{Hyperparameter Selection.} The cornerstone of the training methodology for \ours and the \MQT and \MQCNN baseline models is the definition of the training, validation and test sets.
For the worlwide retail dataset the training set consists of the first four years of observations before a year of validation data. Since \ours is a production model the test set is incrementally updated as new data becomes available, the model selection is performed using an online learning approach.

For the hyperparameter selection, we only consider the exploration learning rates and then number of SGD.
See Table~\ref{table:model_parameters} the configuration space along with particular hyperparameters for each dataset.

\paragraph{Computational Resources and Code.} All the architectures were distributively trained using 40 p3dn.24xlarge machines, each with eight NVIDIA V100 GPUs, 768 GB of memory, and 96 CPUs each. The neural networks are implemented on pytorch~\citep{pytorch2019library} and use the  the DeepTSv2 library~\cite{amazon2024deeptsv2}, however the \MQCNN, \MQT baselines have open source implementations available in the GluonTS library~\citep{alexandrov2020package_gluonts}. The GPU cluster is administrated using torch distributed and the Amazon Sagemaker orchestrator. A complete forecast pipeline including training and inference, takes approximately 20 hours to complete.



\clearpage
\section{SPADE Architecture Ablation Study}

\begin{figure}[!ht] 
    \centering
    \subfigure{
    \includegraphics[width=0.45\linewidth]{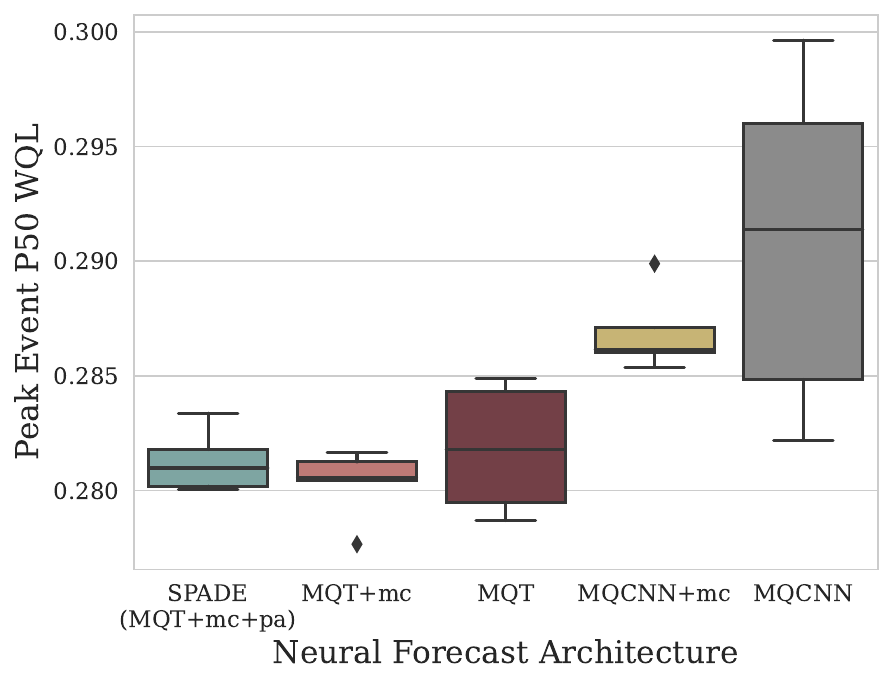} 
    }
    \subfigure{
    \includegraphics[width=0.45\linewidth]{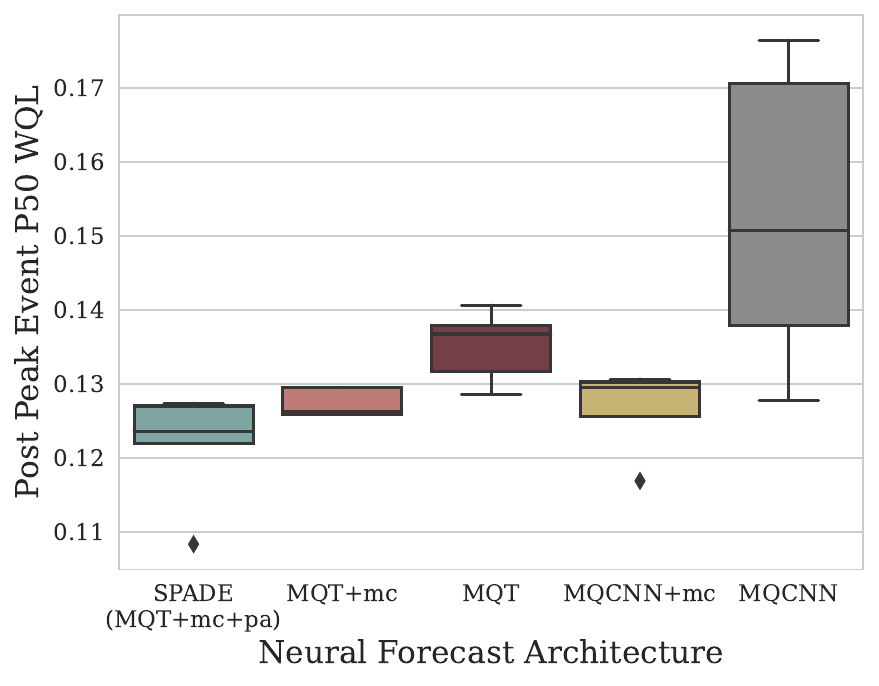} 
    }
    \caption{Peak Attention and Masked Convolutions enhance both MQCNN and MQT architectures by filtering peaks before inputting temporal features to the encoder, thus mitigating the peak carry-over effect. This approach significantly reduces training variance and enhances the forecast accuracy.} 
    \label{fig:ablation_study}
\end{figure}

\begin{table}[ht] 
\tiny
    \begin{center}
    \caption{Evaluation of P50 and P90 probabilistic forecasts averaged over 10 random seeds along their 95 percent confidence interval ($\pm$), lower is better. All hyperparameters were kept constant across all architecture variants. Average percentage difference relative to control in the first column.}
    \label{table:ablation_robust} 
	\begin{tabular}{ll|cccc|ccc} \toprule
    						        &                  & \multicolumn{4}{c}{\MQT}                & \multicolumn{3}{c}{\MQCNN}                \\
    						        &                  & Diff     & Peak Attention & Masked Conv        & Original      & Diff    & Masked Conv         & Original       \\ \midrule
    \parbox[t]{1mm}{\multirow{2}{*}{\rotatebox[origin=c]{0}{Overall}}}
                                    &\;\qquad P50 WQL &    -1.237 & 0.0918±0.0041  & 0.0946±0.0107 &  0.0930±0.0068 &       2.333 &  0.0939±0.0034 &  0.0918±0.0057 \\
                                    &\;\qquad P90 WQL &    -5.786 & 0.0629±0.0007  & 0.0646±0.0029 &  0.0668±0.0020 &      -1.757 &  0.0642±0.0005 &  0.0653±0.0007 \\ \midrule
    \parbox[t]{1mm}{\multirow{2}{*}{\rotatebox[origin=c]{0}{Peak}}}
                                    &\;\qquad P50 WQL &    -0.197 & 0.2813±0.0024  & 0.2803±0.0027 &  0.2818±0.0049 &      -1.342 &  0.2869±0.0031 &  0.2908±0.0129 \\
                                    &\;\qquad P90 WQL &    -4.193 & 0.3483±0.0064  & 0.3422±0.0096 &  0.3636±0.0200 &      -3.736 &  0.3545±0.0104 &  0.3683±0.0323 \\ \midrule
    \parbox[t]{1mm}{\multirow{2}{*}{\rotatebox[origin=c]{0}{PostPeak}}}
                                    &\;\qquad P50 WQL &    -9.966 & 0.1217±0.0137  & 0.1274±0.0033 &  0.1351±0.0086 &     -17.121 &  0.1266±0.0101 &  0.1527±0.0364 \\
                                    &\;\qquad P90 WQL &   -16.151 & 0.0762±0.0038  & 0.0777±0.0017 &  0.0909±0.0051 &     -17.834 &  0.0775±0.0036 &  0.0944±0.0111 \\ \bottomrule
	\end{tabular}
	\end{center}
\end{table}

We performed ablation studies on variants of the \MQCNN\ and \MQT. We attribute the effectiveness of \ours\ in mitigating the peak \emph{carry-over} effect to the incorporation of masked convolutions and temporal feature splitting. We simplified the experimental setup in this ablation study, using the \TourismL dataset, a detailed Australian Tourism Dataset comes from the National Visitor Survey, managed by the Tourism Research Australia agency, it is composed of 555 monthly series from 1998 to 2016 organized geographically~\citep{wickramasuriya2019hierarchical_mint_reconciliation}. We introduce peaks in the time series, simulating a 3\% data contamination using normal noise with the variance of each series to achieve better control in the experiments, making it easier to evaluate the accuracy of around PEs.

We consider the forecasting task, where we produce P50 and P90 quantile forecasts for the last twelve months of all \TourismL series, that we evaluate using the Weighted Quantile Loss (WQL) defined in Equation~\ref{equation:WQL}, but in contrast to the main experiment we only consider weekly forecasts. We differentiate between overall and post PEs (PPEs), to focus on the \emph{carry-over} effect. As shown in Table~\ref{table:ablation_robust}, the masked convolution filters predictions and consistently improves its original counterpart. Relative post PEs P50 improvements of 9.96\% for \MQT and 17.12\% for \MQCNN. Figure~\ref{fig:ablation_study} and Table~\ref{table:ablation_robust}, comparing \MQCNN/\MQT with and without masked convolution reveal a notable enhancement in PPE accuracy when the convolution is included. In addition combining the masked convolutions with the \PeakAttention module produces the best results, as components appear to complement each other. \label{appendix:spade_ablation}

\end{document}